\documentclass[letterpaper]{article} 
\usepackage{aaai23}  
\usepackage{times}  
\usepackage{helvet}  
\usepackage{courier}  
\usepackage[hyphens]{url}  
\usepackage{graphicx} 
\usepackage{natbib}  
\usepackage{caption} 
\usepackage{algorithm}
\usepackage{algorithmic}
\usepackage{amsmath}
\usepackage{amsfonts}
\usepackage{booktabs}

\usepackage{multirow}
\usepackage{xcolor}

\usepackage{newfloat}
\usepackage{listings}
\DeclareCaptionStyle{ruled}{labelfont=normalfont,labelsep=colon,strut=off} 
\floatstyle{ruled}
\newfloat{listing}{tb}{lst}{}
\floatname{listing}{Listing}
\title{High-level semantic feature matters few-shot unsupervised domain adaptation}
\author{
    Lei Yu\textsuperscript{\rm 1}, Wanqi Yang\textsuperscript{\rm 1}\footnote{The corresponding author is Wanqi Yang.}, Shengqi Huang\textsuperscript{\rm 1}, Lei Wang\textsuperscript{\rm 2}, Ming Yang\textsuperscript{\rm 1}\\
}
\affiliations{
    \textsuperscript{\rm 1}School of Computer and Electronic Information, Nanjing Normal University, China\\
    \textsuperscript{\rm 2}School of Computing and Information Technology, University of Wollongong, Australia\\
    yulei@njnu.edu.cn, yangwq@njnu.edu.cn, huangshengqi@njnu.edu.cn, leiw@uow.edu.au, myang@njnu.edu.cn
    
}

\usepackage{bibentry}

\begin{document}

\maketitle

\begin{abstract}
In few-shot unsupervised domain adaptation (FS-UDA), most existing methods followed the few-shot learning (FSL) methods to leverage the low-level local features (learned from conventional convolutional models, \emph{e.g.,} ResNet) for classification. However, the goal of FS-UDA and FSL are relevant yet distinct, since FS-UDA aims to classify the samples in target domain rather than source domain. We found that the local features are insufficient to FS-UDA, which could introduce noise or bias against classification, and not be used to effectively align the domains. To address the above issues, we aim to refine the local features to be more discriminative and relevant to classification. Thus, we propose a novel task-specific semantic feature learning method (TSECS) for FS-UDA. TSECS learns high-level semantic features for image-to-class similarity measurement. Based on the high-level features, we design a cross-domain self-training strategy to leverage the few labeled samples in source domain to build the classifier in target domain. In addition, we minimize the KL divergence of the high-level feature distributions between source and target domains to shorten the distance of the samples between the two domains. Extensive experiments on \textit{DomainNet} show that the proposed method significantly outperforms SOTA methods in FS-UDA by a large margin (\emph{i.e.}, $\sim10\%$).
\end{abstract}

\section{keywords}
Few-shot unsupervised domain adaptation, image-to-class similarity, high-level semantic features, cross-domain self-training, cross-attention.

\section{Introduction}

Currently, a setting namely few-shot unsupervised domain adaptation (FS-UDA) \cite{10.1145/3474085.3475232}\cite{Yang}, which utilizes few labeled data in source domain to train a model to classify unlabeled data in target domain, owns its potential feasibility. 
Typically, a FS-UDA model could learn general knowledge from base classes during training to guide classification in novel classes during testing. 
It is known that both insufficient labels in source domain and large domain shift make FS-UDA as a challenging task.
\begin{figure}[ht]
    \centering
    \setlength{\belowdisplayskip}{-1cm}
    \includegraphics[width=0.9\columnwidth]{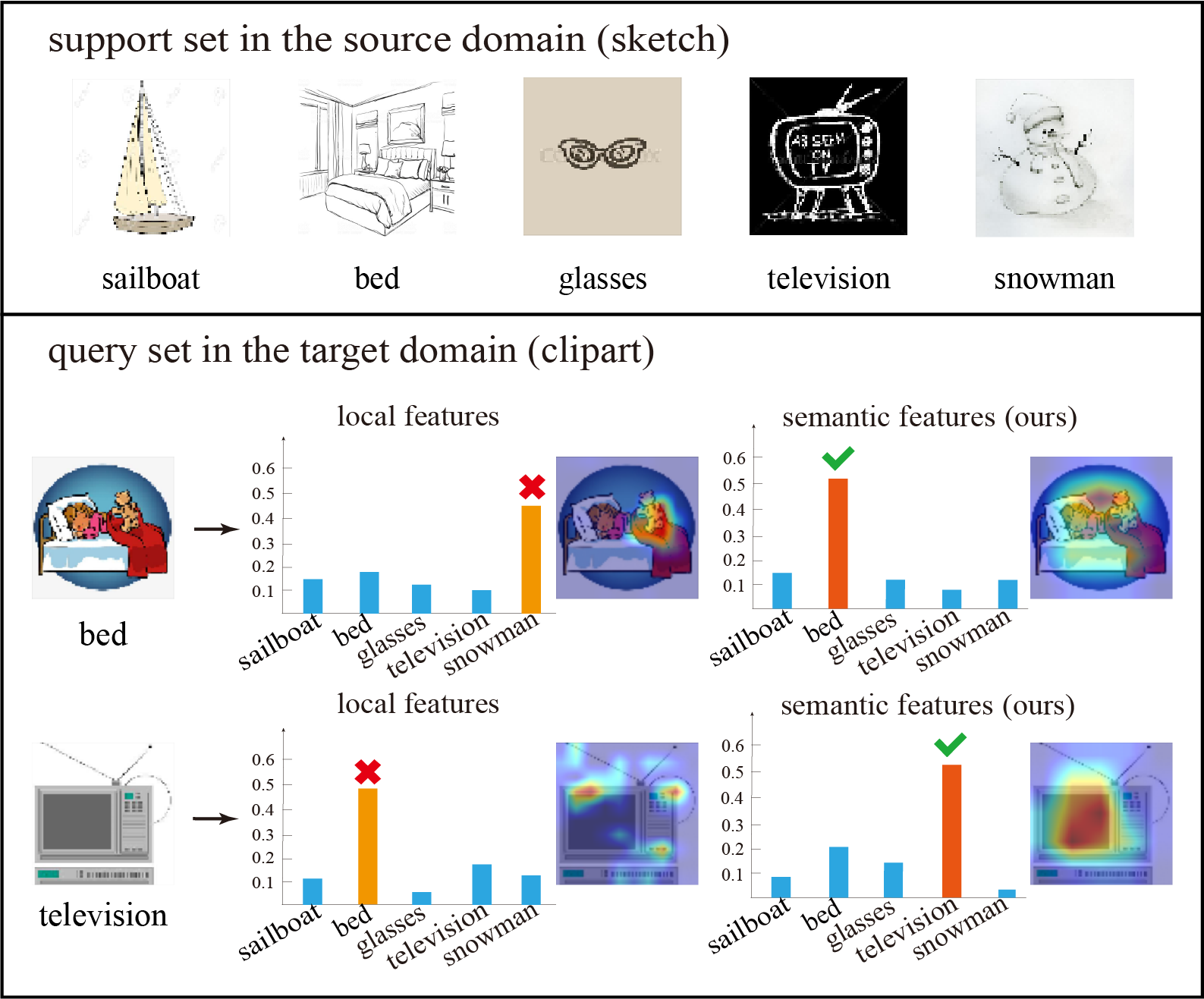}
    \vspace{-0.2cm}
    \caption{A 5-way 1-shot task for FS-UDA where the support set includes five classes and one sample for each class. The figure shows the similarity of query images to every support classes and the spatial similarity of query images to the predicted support class. We found using local features could cause some  inaccurate regions of query images to match the incorrect classes, while our semantic features make the object region in query images similar with their true class, thus achieving correct classification.}
    \label{fig:problem}
    \vspace{-0.4cm}
\end{figure}

Previous studies, \emph{e.g.,} IMSE \cite{10.1145/3474085.3475232}, first followed several few-shot learning (FSL) methods \cite{Li_2019_CVPR}\cite{8099799} to learn the local features by using convolutional models (\emph{e.g.,} ResNet) and then leveraged them to learn image-to-class similarity pattern for classification. However, we wish to clarify that \textit{the goal of FS-UDA and FSL are relevant yet distinct}, since both of them suffer from insufficient labeled training data whereas FS-UDA aims to classify the samples in target domain rather than source domain. 
As shown in Fig. \ref{fig:problem}, by visualizing the spatial similarity of query images to predicted support classes, we found using local features causes the inaccurate regions of query images to match incorrect classes. 
This reason might be that few labeled samples and large domain shift between the support and query sets simultaneously result in the conventional local features in FSL to fail in classification. In this sense, the local features are insufficient to FS-UDA, which could introduce noise or bias against the classification in target domain and not be used to effectively align the domains.

To address this issue, we aim to refine the low-level local features to be more discriminative and relevant to classification, \emph{i.e., high-level semantic features}, and meanwhile align the semantic features for domain adaptation. Therefore, we propose a novel task-specific semantic feature method (TSECS) that learns the semantic features for each task by clustering the local features of support set and query set. To obtain the related semantics from previous tasks, the cluster centroids of the current task are then fused by cross-attention with that of the previous task to generate high-level semantic features to boost classification performance.

\begin{figure}[t]
    \centering
    \includegraphics[width=0.88\columnwidth]{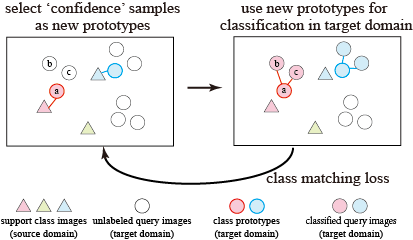}
    \vspace{-0.2cm}
    \caption{Illustration of the process for cross-domain self-training in TSECS. Different shapes represent different domains. We first select the `confidence' target samples (\emph{e.g.,} a) that are very similar to support classes, and then regard them as the new class prototypes to further classify the other target samples (\emph{e.g.,} b, c). This process is executed iteratively with using class matching loss to narrow the distance of query images and their most similar support classes.}
    \label{fig:example}
    \vspace{-0.4cm}
\end{figure}

Moreover, for the domain shift between source and target domains, many domain adaptation methods \cite{Saito_2018_CVPR}\cite{8099799}\cite{DBLP:journals/corr/TzengHZSD14} reduced the distribution discrepancy between domains by using a discriminator to adverse against feature embedding. However, this way could fail in aligning the samples of the same class between domains due to label missing in target domain, which could make the classes of two domains mismatched and thus affect the classification.
Therefore, we aim to align the high-level semantic features by minimizing the KL divergence of the semantic feature distributions between domains, and meanwhile design a cross-domain self-training strategy to train the classifier in target domain. 

We hypothesis that there are usually several `confidence' samples in target domain that could be classified correctly by support set in source domain, in other words, they are very similar to their class prototypes in source domain. Meanwhile, the target domain samples in the same class are more similar to each other than that of other classes. Based on this, we regard these `confidence' samples in the target domain as new prototypes of the classes, which replace those from the support set of source domain.
As shown in Fig. \ref{fig:example}, several `confidence' samples (\emph{e.g.,} a) can be selected as prototypes of their similar classes for classification (\emph{e.g.,} b and c) in target domain. Moreover, the process is conducted iteratively by using class matching loss for better domain alignment.

In sum, we propose the novel method, namely TSECS, for FS-UDA. It refines the local features of convolutional network to generate specific semantic features of each task, and meanwhile perform cross-domain self-training to transport labels from support set in the source domain to query set in the target domain to effectively classify the samples in target domain. Our contributions can be summarized as:
\begin{description}
\item[(1)] \textbf{A novel solution for FS-UDA.} TSECS aims to learn high-level semantic features for classification and domain alignment, which could be regarded as a more effective and efficient way than using local features.
\item[(2)] \textbf{Task-specific semantic embedding for few-shot setting}. It can be seamlessly add to existing FSL/FS-UDA models, which could alleviate the bias of classification. 
\item[(3)] \textbf{Cross-domain self-training for domain alignment.} 
It is designed to bring the samples of the same class close, which could guide effective domain alignment.
\end{description}

We conduct extensive experiments on \textit{DomainNet}. Our method significantly outperforms SOTA methods in FS-UDA by a large margin up to $\sim10\%$.

\section{Related Works}
\textbf{Unsupervised domain adaptation.} The conventional UDA methods aim to reduce discrepancy between source domain and target domain in the feature space and utilize sufficiently labeled source domain data to classify data from target domain. The difference between unsupervised domain adaptation methods often lies in the evaluation of domain discrepancy and the objective function of model training. Several researchers \cite{pmlr-v37-long15}\cite{DBLP:journals/corr/TzengHZSD14} minimize the feature discrepancy by using maximum mean discrepancy to measure the discrepancy between the distribution of domains. Moreover, adversarial training \cite{8099799}\cite{JMLR:v17:15-239} to learn domain-invariant features is usually used to tackle domain shift. Several methods \cite{Tang_2020_CVPR}\cite{9010413}\cite{10.1007/978-3-030-01219-9_18}\cite{9710187}train the classifier in both source domain and target domain and utilize pseudo-labels from target domain to calculate classification loss.
Overall, these UDA methods all require sufficiently labeled source domain data to realize domain alignment and classification, but they perform poor when labeled source domain data are insufficient.

\textbf{Few-shot learning.} Few-shot learning has two main streams, metric-based and optimization-based approaches. Optimization-based methods \cite{bertinetto2018metalearning}\cite{pmlr-v70-finn17a}\cite{ravi2017optimization} usually train a meta learner over auxiliary dataset to learn a general initialization model, which can fine-tune and adapt to new tasks very soon. The main purpose of metric-based methods \cite{Li_2019_CVPR}\cite{NIPS2017_cb8da676}\cite{NIPS2016_90e13578}\cite{Ye_2020_CVPR} is that learn a generalizable feature embedding for metric learning, which can immediately adapt to new tasks without any fine-tune and retraining. Typically, ProtoNet \cite{NIPS2017_cb8da676} learns the class prototypes in the support set and classifies the query images based on the maximum similarity to these prototypes. 
Other than these metric-based methods on feature maps, many methods on local  features
have appeared. DN4 \cite{Li_2019_CVPR} utilizes large amount of local features to measure the similarity between support and query sets instead of flattening the feature map into a long vector. Based on local features, DeepEMD \cite{Zhang_2020_CVPR} adopts Earth Mover's Distance distance to measure the relationship between query and support sets.
Furthermore, a few recent works focus on the issue of cross-domain FSL in which domain shift exists between data of meta tasks and new tasks. The baseline models \cite{chen2018a} are used  to do cross-domain FSL. LFT \cite{Tseng2020Cross-Domain} performs adaptive feature transformation to tackle the domain shift.

\textbf{Few-shot unsupervised domain adaptation.} 
Compared with UDA, FS-UDA is to deal with many UDA tasks by leveraging few labeled source domain samples for each. And compared with cross-domain FSL, FS-UDA are capable of handling the circumstances of no available labels in the target domain, and large domain gap between the support and query sets in every task. For the one-shot UDA \cite{NEURIPS2020_ed265bc9}, it deals with the case that only one unlabeled target sample is available, but does not require the source domain to be few-shot, which is different from ours. Recently, there are a few attempts in FS-UDA. PCS \cite{Yue_2021_CVPR} performs prototype self-supervised learning in cross-domain, but they require enough unlabeled source samples to learn prototypes and ignore task-level transfer, which is also different from ours. meta-FUDA \cite{Yang} leverages meta learning-based optimization to perform task-level transfer and domain-level transfer jointly. IMSE \cite{10.1145/3474085.3475232} utilizes local features to learn similarity patterns for cross-domain similarity measurement. However, they did not consider that local features could bring the noise or bias to affect classification and domain alignment. Thus, we propose task-specific semantic features to solve this problem.

\section{Methodology}
\subsection{Problem Definition}

\textbf{A $N$-way, $K$-shot FS-UDA task.} Table \ref{tab:Notations} shows the main symbols used in this paper. The FS-UDA setting includes two domains: a source domain $S$ and a target domain $T$. A $N$-way, $K$-shot FS-UDA task includes a support set $X_S$ from $S$ and a query set $Q_T$ from $T$. The support set $X_S$ contains $N$ classes and $K$ samples per class in the source domain. The query set $Q_T$ contains the same $N$ classes as in $X_S$ and $N_q$ target domain samples per class. To classify query images in $Q_T$ to the correct class in $X_S$, 
it is popular to train a general model from base classes to adapt to handle new $N$-way, $K$-shot FS-UDA tasks for testing.

\begin{table}
\caption{Notations}
\vspace{-0.4cm}
\label{tab:Notations}
\centering
\renewcommand{\arraystretch}{1}
\setlength{\tabcolsep}{3pt}
\scalebox{0.77}{
\begin{tabular}{c|c}
\toprule[2pt]
\textbf{Notations} & \textbf{Descriptions} \\
\hline
$N \in \mathbb{R}$ & The number of classes in the task.\\
\hline
$K \in \mathbb{R}$ & The number of samples per class in support set.\\
\hline
\multirow{2}{*}{$X_S, Q_S, Q_T$} & Support set of source domain, and query sets \\
&of source domain and target domain.\\
\hline
$H, W, d \in \mathbb{R}$ & The height, width, and channel of feature map.\\
\hline
$L \in \mathbb{R}^{HW \times d}$ & The local feature vectors in the feature map.\\
\hline
$k \in \mathbb{R}$ & The number of semantic clusters for an episode.\\
\hline
$C \in \mathbb{R}^{k \times d}$ & The centroids of the clusters.\\
\hline
$F,\hat{F},$ & The semantic feature map, semantic features and\\
$\hat{F}_{X_S},\hat{F}_{Q_S},\hat{F}_{Q_T}$& the parts of support and query sets in both domains.\\
\hline
$M^c_q \in \mathbb{R}^{H \times W \times N}$ & The 3-D similarity matrix for classification.\\
\hline
$p^c_q \in \mathbb{R}^{KHW}$ & Similarity pattern vectors of a query image $q$ \\
$p^i_q \in \mathbb{R}^{HW}$ & with a support class $c$ and a support image $i$,\\
$p^{pos}_q, p^{neg}_q\in \mathbb{R}^{KHW}$ & and the most similar class and the second one for $q$.\\
\hline
$\mu_A, \mu_B \in \mathbb{R}^{HW \times d}$& The mean of semantic features or similarity patterns.\\
\hline
\multirow{2}{*}{$\Sigma_A, \Sigma_B \in \mathbb{R}^{HW \times HW}$} & Covariance matrix of semantic features\\ 
& or similarity patterns.\\
\hline
$\lambda_{sfa}, \lambda_{spa}, \lambda_{clm}$ & Weight parameters of three loss terms in Eq. (\ref{total loss}).\\
\bottomrule[2pt]
\end{tabular}
}
\vspace{-0.4cm}
\end{table}

\textbf{Auxiliary dataset and episodic training.} As in \cite{10.1145/3474085.3475232}, the base classes are collected from an auxiliary dataset $D^{aux}$ to perform episodic training to learn the general model. Note that the base classes in $D^{aux}$ are completely different from new classes in testing tasks, which are unseen during episodic training. Moreover, $D^{aux}$ includes labeled source domain data and unlabeled target domain data for FS-UDA. 
We construct large amounts of episodes, each containing $\{X_S,Q_S,Q_T\}$ as in \cite{10.1145/3474085.3475232}, to simulate the testing tasks for task-level generalization. Note that $Q_S$ is introduced into episodic training to calculate classification loss and perform domain alignment with $Q_T$.

\begin{figure*}
    \centering
    \includegraphics[width=0.9\textwidth]{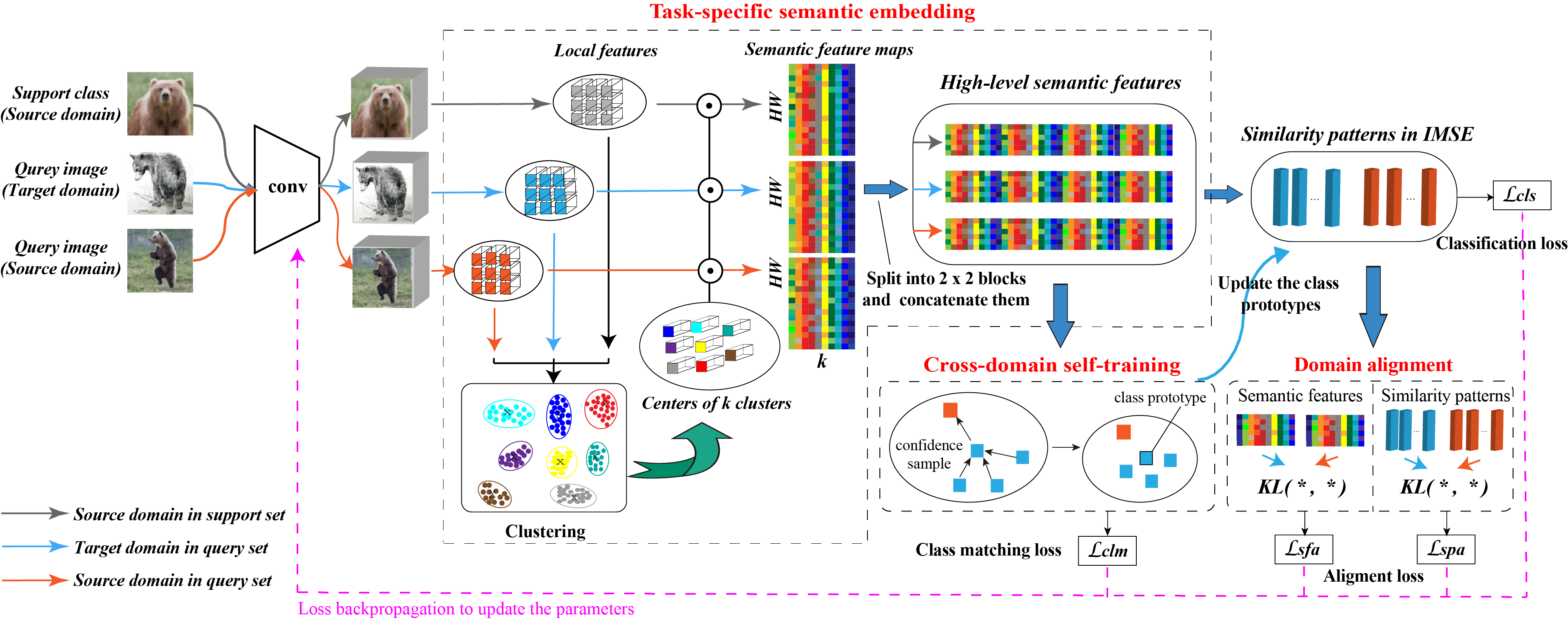}
    \vspace{-0.3cm}
    \caption{Illustration of our method training per episode for 1-shot FS-UDA tasks. 
    First, support classes and query images from both domains are through a convolution network to extract their local features, followed by the task-specific semantic embedding module to learn high-level semantic features. Then, these semantic features are fed into the cross-domain self-training module to update the class prototypes for target domain classification and calculate the class matching loss $\mathcal{L}_{clm}$.
    Meanwhile, these semantic features are also used to generate similarity patterns in IMSE \cite{10.1145/3474085.3475232} for classification loss $\mathcal{L}_{cls}$. In addition, both semantic features and similarity patterns from both domains are aligned by the domain alignment module with the alignment losses $\mathcal{L}_{sfa}$ and $\mathcal{L}_{spa}$, respectively. Finally, all the losses are backpropagated to update our model.}
    \label{fig:overview}
    \vspace{-0.3cm}
\end{figure*}

\textbf{The flowchart of our method.} Fig. \ref{fig:overview} illustrates our method for 5-way, 1-shot FS-UDA tasks. In each episode, a support set ($X_S$) and two query sets ($Q_S$ and $Q_T$) are first through the convolution network (\emph{e.g., ResNet}) to extract their local features. Then, the task-specific semantic embedding module refines the local features to generate semantic features, which is computational efficient due to dimension reduction.
Also, based on semantic features of $Q_S$ and $Q_T$, we leverage their similarity patterns \cite{10.1145/3474085.3475232} to calculate \emph{image-to-class} similarity for classification with the loss $\mathcal{L}_{cls}$. To improve its performance, cross-domain self-training module is performed to introduce the class prototypes of target domain and train a target domain classifier with a class matching loss $\mathcal{L}_{clm}$. In addition, the semantic features and similarity patterns from both domains are further aligned by calculating their alignment losses $\mathcal{L}_{sfa}$ and $\mathcal{L}_{spa}$, respectively. Finally, the losses above are back-propagated to update our model. After episodic training over all episodes, we utilize the learned model to test new FS-UDA tasks. 
Then, we calculate the averaged classification accuracy on these tasks for performance evaluation.

\subsection{\textbf{Task-specific Semantic Feature Learning}}

Most FSL methods and FS-UDA methods learned local features from convolutional networks for classification. However, we found that the local features could introduce noise or bias that is valid for classification and domain alignment. Thus, we aim to refine the local features to generate high-level semantic features for each task. In the following, we will introduce our semantic feature embedding module. 

First of all, in each episode, all local features $L\in\mathbb{R}^{(|X_S|+|Q_S|+|Q_T|)HW \times d}$ are extracted from the convolutional network, where $|\cdot|$ is the number of samples in a set.
Then, we cluster the local features to generate different semantic clusters for support set and query set, respectively, since clustering the two sets together could result in the clusters that relate to the domains due to the presence of large domain gap.
For simplification, we adopt K-means for clustering, and meanwhile utilize the singular value decomposition (SVD) to adaptively take the number of eigenvalues greater than a certain threshold as the cluster number $k$ ($k\ll d$) for each task. Afterwards, we calculate the task-specific semantic feature map $F\in\mathbb{R}^{(|X_S|+|Q_S|+|Q_T|)HW \times k}$ by measuring the \emph{Cosine} similarity between the local features $L$ and the centroids $C\in\mathbb{R}^{k \times d}$ 
of all semantic clusters, \emph{i.e.,} 
$F = \frac{L}{{||L||}_2} \cdot \frac{C^{\top}}{{||C||}_2}$. 
Finally, we split $F$ to $2\times2$ blocks based on height and weight dimension of the feature map, and then concatenate the four blocks together along the channel to generate semantic features $\hat{F}\in\mathbb{R}^{\frac{1}{4}(|X_S|+|Q_S|+|Q_T|)HW \times 4k}$. 
This is a simple yet effective way to maintain discriminative ability and spatial information of semantic features. 

Moreover, to leverage the semantics from previous tasks to guide the semantic feature learning of the current task, we utilize the centroids of previous clusters to update the initialization of clustering centroids by cross-attention \cite{9711309}. This makes K-means clustering converge rapidly.

After obtaining the semantic features $\hat{F}$, we use them for domain alignment and classification. Firstly, $\hat{F}$ is partitioned into $\hat{F}_{X_S}$, $\hat{F}_{Q_S}$, $\hat{F}_{Q_T}$ along with the first dimension. Then, we align $\hat{F}_{Q_S}$ and $\hat{F}_{Q_T}$ by minimizing the KL divergence of their distributions that will be introduced later. Meanwhile, we utilize $\hat{F}_{X_S}$, $\hat{F}_{Q_S}$ and  $\hat{F}_{Q_T}$ to build 3-D similarity matrix $M_q^c$ \cite{10.1145/3474085.3475232} between support and query sets. Finally, we calculate the similarity pattern $p_q^c$ (measuring the similarity between query sample $q$ and support class $c$) for classification \cite{10.1145/3474085.3475232}. The classification loss using cross-entropy can be written by:
\begin{equation}
\label{cls loss}
\setlength\abovedisplayskip{0.5pt}
\setlength\belowdisplayskip{0.5pt}
\mathcal{L}_{cls} = -\frac{1}{|Q_S|}\sum_{q \in Q_S}\log(\frac{\exp(\textbf{1} \cdot p^{c}_q)}{\sum_{i=1}^K \exp(\textbf{1} \cdot p^{i}_q)})
\end{equation}

\subsection{\textbf{Cross-domain Self-training}}
Since there is large domain shift between source and target domains, as well as label missing in target domain, adversarial domain adaptation on low-level local features cannot make samples of the same class between domains close, and thus could make the classes of two domains mismatched.

To alleviate the mismatching issue, we aim to find the most similar `confidence' samples in $Q_T$ with $X_S$ to guide classification in target domain. We assume that it usually exists that the `confidence' samples in $Q_T$ could be classified correctly by $X_S$, when the distributions between domains are aligned. We iteratively select the `confidence' samples in $Q_T$ as the new prototypes to replace that in $X_S$ for classification, as shown in Fig. \ref{fig:example}. We call the process as \emph{cross-domain self-training}. The process can find more `confidence' samples from $Q_T$ than that in $X_S$ for the same class, which could correct some misclassified samples in $Q_T$, thereby lightening the impact of domain gap.

Moreover, to improve the performance of the target domain classifier, we aim to make target domain samples $q$ in $Q_T$ closer to their most similar class and meanwhile far away from the other classes. Thus, we first calculate its similarity patterns $p_q^{pos}$ (with the most similar class) and  $p_q^{neg}$ (with the second similar class), and then design the class matching loss with a margin $m$, which can be written by
\begin{equation}
\label{clm loss}
\mathcal{L}_{clm}=\sum_{q \in Q_T}\max(\text{softmax}(p_q^{neg}) -
\text{softmax}(p_q^{pos}) + m, 0),
\end{equation}
where the similarity to the most similar class should be greater by $m$ than the second similar class. 

\subsection{\textbf{Two-level Domain Alignment}}
Conventional adversarial domain adaptation methods \cite{JMLR:v17:15-239}\cite{8099799} iteratively train a discriminator to align the distribution of domains by adversarial training among tasks. However, they cannot be used to align the semantic features, because our semantic features are relevant to tasks, the semantics of the same channel could be varied for different tasks.
Meanwhile, symmetrical alignment could bring the inference information of the target domain to the source domain \cite{ijcai2020-409}. Thus, we use asymmetrical KL divergence to align the distribution of domains on both semantic features and similarity patterns within a task. Then, KL divergence can be calculated by:
\begin{equation}
\label{KL divergence}
\setlength\abovedisplayskip{1.5pt}
\begin{split}
KL(A,B) = & \frac{1}{2}\big(tr(\Sigma_{A}^{\text{-1}}\Sigma_{B}) + \ln(\frac{\Sigma_{A}}{\Sigma_{B}}) \\+&(\mu_{A}-
 \mu_{B})\Sigma_{A}^{\text{-1}}(\mu_{A}-\mu_{B})^{\top}-d\big),
\end{split}
\end{equation}
where $\mu_{A}, \mu_{B}, \Sigma_A$ and $\Sigma_B$ are the mean vectors and the covariance matrices of sample matrix $A$ and $B$, respectively. Thus, we minimize the KL divergence between semantic features $\hat{H}_{Q_S}$ and $\hat{H}_{Q_T}$ by
\begin{equation}
\label{sfa loss}
\setlength\abovedisplayskip{1.5pt}
\setlength\belowdisplayskip{1.5pt}
\mathcal{L}_{sfa} = KL(\hat{F}_{Q_S}, \hat{F}_{Q_T}).
\end{equation}
Meanwhile, we also minimize the KL divergence to align the similarity patterns $\{p^c_{q_S}\}$ of $Q_S$ and $\{p^c_{q_T}\}$ of $Q_T$ with class $c$, which can be written by
\begin{equation}
\label{spa loss}
\setlength\abovedisplayskip{0.5pt}
\setlength\belowdisplayskip{1.5pt}
\mathcal{L}_{spa} 
=
\sum^N_{c=1}KL(\{p^c_{q_S}\},\{p^c_{q_T}\}).
\end{equation}

In sum, we combine all the above losses, w.r.t. classification (Eq. (\ref{cls loss})), class matching (Eq. (\ref{clm loss})) and KL-based domain alignment (Eqs. (\ref{sfa loss}) and (\ref{spa loss})) to train our model on many episodes. The total objective function can be written by:
\begin{equation}
\label{total loss}
\min \mathcal{L}_{cls}+\lambda_{sfa}\mathcal{L}_{sfa}+\lambda_{spa}\mathcal{L}_{spa}+\lambda_{clm}\mathcal{L}_{clm},
\end{equation}
where the hyper-parameters $\lambda_{sfa}$, $\lambda_{spa}$ and $\lambda_{clm}$ are introduced to balance the effect of different loss terms.

\section{Experiment}

\textbf{\emph{DomainNet} dataset.} We conduct extensive experiments on a multi-domain benchmark dataset \emph{DomainNet} to demonstrate the efficacy of our method. It was released in 2019 for the research of multi-source domain adaptation \cite{9010750}. It contains 345 categories and six domains per category, \emph{i.e.,} \emph{quickdraw}, \emph{clipart}, \emph{real}, \emph{sketch}, \emph{painting} and \emph{infograph} domains. In our experiments, we follow the setting of IMSE in \cite{10.1145/3474085.3475232} to remove data insufficient domain \emph{infograph}. There are 20 combinations totally for evaluation, and the dataset is split into 217, 43 and 48 categories for episodic training, model validation and testing new tasks, respectively. Note that in each split every category contains the five-domain images.

\textbf{Network architecture and setting.} We employ ResNet-12 as the backbone of feature embedding network, which is widely used in few-shot learning \cite{10.1145/3474085.3475232} \cite{Gidaris_2020_CVPR}.  
We obtain semantic features by first clustering the local features from each class of support set and two query sets and then concatenating them. During this process, we adopt cross-attention 
that consists of three convolution parameters to generate $(Q,K,V)$ for attention calculation. In cross-domain self-training module, we set the threshold 1.7 of similarity score to select the `confidence' samples in target domain. The margin $m$ in Eq. (\ref{clm loss}) is empirically set to 1.5. In addition, we follow the setting of IMSE \cite{10.1145/3474085.3475232} to obtain similarity patterns. The hyper-parameters $\lambda_{sfa}$, $\lambda_{spa}$ and $\lambda_{clm}$ are set to 0.1, 0.05 and 0.01, by grid search, respectively.

\textbf{Model training, validation and testing.}  To improve the performance, before episodic training, the feature embedding network is pretrained by using source domain data in the auxiliary dataset, as in \cite{10.1145/3474085.3475232}. Afterwards, we perform episodic training on 280 episodes, following the setting of \cite{10.1145/3474085.3475232}. During episode training, the total loss in Eq. (\ref{total loss}) is minimized to optimize the network parameters for each episode. Also, we employ Adam optimizer with an initial learning rate of $10^{\text{-}4}$, and meanwhile reduce the learning rate by half every 280 episodes. For model validation, we compare the performance of different model parameters on $100$ tasks, which is randomly sampled from the validate set containing 43 categories. Then, we select the model parameters with the best validation accuracy for testing. During the testing, we 
randomly select $3000$ tasks to calculate the averaged top-1 accuracy on these tasks as the evaluation criterion.

\begin{table*}[h]
\caption{Comparison of our method with the related methods for 5-way 1-shot or 5-shot FS-UDA tasks. The first three blocks and IMSE are reported from \cite{10.1145/3474085.3475232}, while the last two are the variant of IMSE we designed and ours, respectively. Each row represents the accuracy (\%) of a compared method adapting between two domains, where the \emph{skt, rel, qdr, pnt}, and \emph{cli} denote the \emph{sketch, real, quickdraw, painting}, and \emph{clipart} domains in \emph{DomainNet}, respectively. The best results are in bold.}\label{tab:compare}
\vspace{-0.4cm}
\centering
\renewcommand{\arraystretch}{1.1}
\scalebox{0.67}{
\begin{tabular}{c|cccccccccc|c}
\toprule[2pt]
\multicolumn{12}{c}{5-way, 1-shot}\\
\hline
\multirow{2}{*}{\textbf{Methods}} &
$skt \longleftrightarrow rel$ & $skt \longleftrightarrow qdr$ & $skt \longleftrightarrow pnt$ & $skt \longleftrightarrow cli$ & $rel \longleftrightarrow qdr$ & $rel \longleftrightarrow pnt$ & $rel \longleftrightarrow cli$ & $qdr \longleftrightarrow pnt$ & $qdr \longleftrightarrow cli$ & $pnt \longleftrightarrow cli$ & avg\\
& $\rightarrow/\leftarrow$ & $\rightarrow/\leftarrow$ & $\rightarrow/\leftarrow$ & $\rightarrow/\leftarrow$ & $\rightarrow/\leftarrow$ & $\rightarrow/\leftarrow$ & $\rightarrow/\leftarrow$ & $\rightarrow/\leftarrow$ & $\rightarrow/\leftarrow$ & $\rightarrow/\leftarrow$ & - \\
\hline
\textbf{MCD} & 48.07/37.74   & 38.90/34.51   & 39.31/35.59    & 51.43/38.98  &  24.17/29.85  & 43.36/47.32   & 44.71/45.68 & 26.14/25.02 &  42.00/34.69 & 39.49/37.28  & 38.21  \\
\textbf{ADDA} &  48.82/46.06  &  38.42/40.43  &  42.52/39.88  & 50.67/47.16 &  31.78/35.47  &  43.93/45.51  &  46.30/47.66  & 26.57/27.46 & 46.51/32.19 &  39.76/41.24 &  40.91  \\
\textbf{DWT}   & 49.43/38.67 & 40.94/38.00 & 44.73/39.24 & 52.02/50.69 & 29.82/29.99 & 45.81/50.10 & 52.43/51.55 & 24.33/25.90 & 41.47/39.56 & 42.55/40.52 & 41.38 \\
\hline
\textbf{ProtoNet} & 50.48/43.15 & 41.20/32.63 & 46.33/39.69 & 53.45/48.17 & 32.48/25.06 & 49.06/50.30 & 49.98/51.95 & 22.55/28.76 & 36.93/40.98 & 40.13/41.10 & 41.21 \\
\textbf{DN4} & 52.42/47.29 & 41.46/35.24 & 46.64/46.55 & 54.10/51.25 & 33.41/27.48 & 52.90/53.24 & 53.84/52.84 & 22.82/29.11 & 36.88/43.61 & 47.42/43.81 & 43.61\\
\textbf{ADM} & 49.36/42.27 & 40.45/30.14 & 42.62/36.93 & 51.34/46.64 & 32.77/24.30 & 45.13/51.37 & 46.80/50.15 & 21.43/30.12 & 35.64/43.33 & 41.49/40.02 & 40.11\\
\textbf{FEAT} & 51.72/45.66 & 40.29/35.45 & 47.09/42.99 & 53.69/50.59 & 33.81/27.58 & 52.74/53.82 & 53.21/53.31 & 23.10/29.39 & 37.27/42.54 & 44.15/44.49 & 43.14\\
\textbf{DeepEMD} & 52.24/46.84 & 42.12/34.77 & 46.64/43.89 & 55.10/49.56 & 34.28/28.02 & 52.73/53.26 & 54.25/54.91 & 22.86/28.79 & 37.65/42.92 & 44.11/44.38 & 43.46\\
\hline
\textbf{ADDA+ProtoNet}  & 51.30/43.43 & 41.79/35.40 & 46.02/41.40 & 52.68/48.91 & 37.28/27.68 & 50.04/49.68 & 49.83/52.58 & 23.72/32.03 & 38.54/44.14 & 41.06/41.59 & 42.45 \\
\textbf{ADDA+DN4} & 53.04/46.08 & 42.64/36.46 & 46.38/47.08 & 54.97/51.28 & 34.80/29.84 & 53.09/54.05 & 54.81/55.08 & 23.67/31.62 & 42.24/45.24 & 46.25/44.40 & 44.65 \\
\textbf{ADDA+ADM} & 51.87/45.08 & 43.91/32.38 & 47.48/43.37 & 54.81/51.14 & 35.86/28.15 & 48.88/51.61 & 49.95/54.29 & 23.95/33.30 & 43.59/48.21 & 43.52/43.83 & 43.76 \\
\textbf{ADDA+FEAT} & 52.72/46.08 & 47.00/36.94 & 47.77/45.01 & 56.77/52.10 & 36.32/30.50 & 49.14/52.36 & 52.91/53.86 & 24.76/35.38 & 44.66/48.82 & 45.03/45.92 & 45.20 \\
\textbf{ADDA+DeepEMD} & 53.98/47.55 & 44.64/36.19 & 46.29/45.14 & 55.93/50.45 & 37.47/30.14 & 52.21/53.32 & 54.86/54.80 & 23.46/32.89 & 39.06/46.76 & 45.39/44.65 & 44.75 \\
\hline
\textbf{IMSE}& 57.21/51.30 & 49.71/40.91 & 50.36/46.35 & 59.44/54.06 & 44.43/36.55 & 52.98/55.06  & 57.09/57.98  & 30.73/38.70 &  48.94/51.47 &  47.42/46.52 & 48.86\\
\textbf{IMSE+TSE} & 60.71/56.15 & 53.78/48.57 & 56.50/48.59 & 61.59/56.59 & 45.48/49.45 & 55.44/57.45 & 59.60/59.52 & 37.94/39.83 & 58.83/56.22 & 49.19/51.01 & 52.79\\
\hline
\textbf{TSECS (ours)} & \textbf{65.00/58.22} & \textbf{62.25/51.97} & \textbf{56.51/53.70} & \textbf{69.45/64.59} & \textbf{56.66/49.82} & \textbf{58.76/63.18} & \textbf{67.98/67.89} & \textbf{38.26/46.15} & \textbf{60.51/69.03} & \textbf{54.40/52.76} & \textbf{58.20} \\
\midrule[2pt]
\multicolumn{12}{c}{5-way, 5-shot}\\
\hline
\multirow{2}{*}{\textbf{Methods}} &
$skt \longleftrightarrow rel$ & $skt \longleftrightarrow qdr$ & $skt \longleftrightarrow pnt$ & $skt \longleftrightarrow cli$ & $rel \longleftrightarrow qdr$ & $rel \longleftrightarrow pnt$ & $rel \longleftrightarrow cli$ & $qdr \longleftrightarrow pnt$ & $qdr \longleftrightarrow cli$ & $pnt \longleftrightarrow cli$ & avg\\
& $\rightarrow/\leftarrow$ & $\rightarrow/\leftarrow$ & $\rightarrow/\leftarrow$ & $\rightarrow/\leftarrow$ &
$\rightarrow/\leftarrow$ & $\rightarrow/\leftarrow$ & $\rightarrow/\leftarrow$ & $\rightarrow/\leftarrow$ & $\rightarrow/\leftarrow$ & $\rightarrow/\leftarrow$ & - \\
\hline
\textbf{MCD}  & 66.42/47.73 & 51.84/39.73 & 54.63/47.75 & 72.17/53.23 & 28.02/33.98 & 55.74/66.43 & 56.80/63.07 & 28.71/29.17 & 50.46/45.02 & 53.99/48.24 & 49.65 \\
\textbf{ADDA} & 66.46/56.66 & 51.37/42.33 & 56.61/53.95 & 69.57/65.81 & 35.94/36.87 & 58.11/63.56 & 59.16/65.77 & 23.16/33.50 & 41.94/43.40 & 55.21/55.86 & 51.76 \\
\textbf{DWT}& 67.75/54.85 & 48.59/40.98 & 55.40/50.64 & 69.87/59.33 & 36.19/36.45 & 60.26/68.72 & 62.92/67.28 & 22.64/32.34 & 47.88/50.47 & 49.76/52.52 & 51.74 \\\hline
\textbf{ProtoNet} & 65.07/56.21 & 52.65/39.75 & 55.13/52.77 & 65.43/62.62 & 37.77/31.01 & 61.73/66.85 & 63.52/66.45 & 20.74/30.55 & 45.49/55.86 & 53.60/52.92 & 51.80 \\
\textbf{DN4} & 63.89/51.96 & 48.23/38.68 & 52.57/51.62 & 62.88/58.33 & 37.25/29.56 & 58.03/64.72 & 61.10/62.25 & 23.86/33.03 & 41.77/49.46 & 50.63/48.56 & 49.41 \\
\textbf{ADM}  & 66.25/54.20 & 53.15/35.69 & 57.39/55.60 & 71.73/63.42 & 44.61/24.83 & 59.48/69.17 & 62.54/67.39 & 21.13/38.83 & 42.74/58.36 & 56.34/52.83 & 52.78 \\
\textbf{FEAT}  & 67.91/58.56 & 52.27/40.97 & 59.01/55.44 & 69.37/65.95 & 40.71/28.65 & 63.85/71.25 & 65.76/68.96 & 23.73/34.02 & 42.84/53.56 & 57.95/54.84 & 53.78 \\
\textbf{DeepEMD}  & 67.96/58.11 & 53.34/39.70 & 59.31/56.60 & 70.56/64.60 & 39.70/29.95 & 62.99/70.93 & 65.07/69.06 & 23.86/34.34 & 45.48/53.93 & 57.60/55.61 & 53.93 \\
\hline
\textbf{ADDA+ProtoNet} & 66.11/58.72 & 52.92/43.60 & 57.23/53.90 & 68.44/61.84 & 45.59/38.77 & 60.94/69.47 & 66.30/66.10 & 25.45/41.30 & 46.67/56.22 & 58.20/52.65 & 54.52 \\
\textbf{ADDA+DN4} & 63.40/52.40 & 48.37/40.12 & 53.51/49.69 & 64.93/58.39 & 36.92/31.03 & 57.08/65.92 & 60.74/63.13 & 25.36/34.23 & 48.52/51.19 & 52.16/49.62 & 50.33 \\
\textbf{ADDA+ADM} & 64.64/54.65 & 52.56/33.42 & 56.33/54.85 & 70.70/63.57 & 39.93/27.17 & 58.63/68.70 & 61.96/67.29 & 21.91/39.12 & 41.96/59.03 & 55.57/53.39 & 52.27 \\
\textbf{ADDA+FEAT} & 67.80/56.71 & 60.33/43.34 & 57.32/58.08 & 70.06/64.57 & 44.13/35.62 & 62.09/70.32 & 57.46/67.77 & 29.08/44.15 & 49.62/63.38 & 57.34/52.13 & 55.56 \\
\textbf{ADDA+DeepEMD} & 68.52/59.28 & 56.78/40.03 & 58.18/57.86 & 70.83/65.39 & 42.63/32.18 & 63.82/71.54 & 66.51/69.21 & 26.89/42.33 & 47.00/57.92 & 57.81/55.23 & 55.49 \\
\hline
\textbf{IMSE}  & 70.46/61.09 & 61.57/46.86  & 62.30/59.15 & 76.13/67.27 & 53.07/40.17 & 64.41/70.63 & 67.60/71.76 & 33.44/48.89 & 53.38/65.90 & 61.28/56.74 & 59.60\\
\textbf{IMSE+TSE} & 72.75/62.24 & 64.49/55.04 & 62.86/61.10 & 77.39/69.87 & 53.88/54.48 & 63.97/72.46 & 69.86/72.49 & 37.43/51.66 & 64.43/67.46 & 63.40/57.89 & 62.76\\
\hline
\textbf{TSECS (ours)} & \textbf{78.23/70.44} & \textbf{77.90/55.77} & \textbf{66.70/68.03} & \textbf{83.82/74.28} & \textbf{64.33/55.16} & \textbf{68.40/79.74} & \textbf{78.23/77.69} & \textbf{39.74/63.02} & \textbf{67.99/80.31} & \textbf{73.67/61.63} & \textbf{69.25}\\

\bottomrule[2pt]
\end{tabular}
}
\vspace{-0.45cm}
\end{table*}

\subsection{Comparison Experiments for FS-UDA}
We conduct extensive experiments on \emph{DomainNet} to compare our method with five FSL methods (ProtoNet \cite{NIPS2017_cb8da676}, DN4 \cite{Li_2019_CVPR}, ADM \cite{ijcai2020-409}, FEAT \cite{Ye_2020_CVPR}, DeepEMD \cite{Zhang_2020_CVPR}), three UDA methods, (MCD \cite{Saito_2018_CVPR}, ADDA \cite{8099799}, DWT \cite{Roy_2019_CVPR}), their combinations and the most related method IMSE \cite{10.1145/3474085.3475232}. For fair comparison, the results of these above methods are all reported from \cite{10.1145/3474085.3475232} with the same setting. Moreover, we also modify IMSE by using our semantic features for classification and domain adversary, namely IMSE+TSE. For fair comparison, these compared methods also pretrain the embedding network before episodic training, and they are trained on 1000 episodes. 

\textbf{Comparison analysis.} Table \ref{tab:compare} shows the results of all the compared methods for 20 cross-domain combinations, which records the averaged classification accuracy of target domain samples over 3000 5-way 1-shot/5-shot FS-UDA tasks. As observed, our TSECS achieves the best performance for all combinations and their average. Specifically, the UDA and  FSL baselines in the first two parts perform the worst. 
In the third part, the combination methods with ADDA \cite{8099799} perform domain adversarial training each episode, thus generally better than the above two parts, but still inferior to IMSE \cite{10.1145/3474085.3475232} and our TSECS. This is because the combination methods only perform domain alignment based on original feature maps, not considering the alignment of similarity patterns (related to classification predictions). Also, IMSE is worse than IMSE+TSE, which indicates high-level semantic features are more effective for FS-UDA than local features. However, they are still much worse than our method, showing the efficacy of high-level semantic features and cross-domain self-training for FS-UDA.

On the other hand, we can see that the 20 cross-domain combinations have considerably different performances. This is because several domains (\emph{e.g., quickdraw}) are significantly different from other domains, while several other domains (e.g. \emph{real, clipart}) are with the similar styles and features. Thus, for most compared methods, the performance becomes relatively low when the domain gap is large.
For example, from \emph{quickdraw} to \emph{painting}, it performs the worst in all the other combinations because of larger domain gap, but our TSECS outperforms IMSE and the other compared methods by $8\%$ and $12\%$, respectively. We found that our method has the larger performance improvement over IMSE, for these combinations containing \emph{quickdraw}, which shows the efficacy of our method for large domain gap. Also, like TSECS, IMSE+TSE performs much better than IMSE for large domain gap, which indicates the high-level semantic features could conduct domain adaptation better than local features. In sum, these results reflect the advantages of our TSECS to deal with domain shift and task generalization in FS-UDA, no matter how large the domain gap is.

\begin{table}
\caption{Ablation study  ($\%$) of the modules designed in TSECS, where
the FS-UDA tasks are evaluated from a domain (\emph{sketch}) to the other four domains in \emph{DomainNet}.}
\label{tab:ablation module}
\vspace{-0.4cm}
\centering
\scriptsize
\setlength{\tabcolsep}{10pt}
\renewcommand{\arraystretch}{0.95}
\begin{tabular}{ccc|cccc}
\toprule[1.5pt]
\multicolumn{3}{c|}{\textbf{Components}} & \multicolumn{4}{c}{\textbf{Target Domains}} \\
\hline
\emph{TSE} &\emph{catt}& \emph{CS}  & \emph{cli} & \emph{rel} & \emph{qdr} & \emph{pnt} \\
\hline
$\checkmark$ & & & 61.98 & 60.00 & 52.21 & 51.62 \\
& & \checkmark & 57.07 & 53.31 & 41.93 & 46.66\\
$\checkmark$ & $\checkmark$ & & 62.74 & 60.54 & 53.64 & 54.23 \\ 
$\checkmark$ & & $\checkmark$ & 68.25 & 61.15 & 58.31 & 53.34\\
$\checkmark$ & $\checkmark$ & $\checkmark$ & \textbf{69.45} & \textbf{65.00} & \textbf{62.25} & \textbf{56.51}\\
\bottomrule[1.5pt]
\end{tabular}
\vspace{-0.5cm}
\end{table}

\textbf{Ablation study of our method.} We conduct various experiments on \emph{DomainNet} to evaluate the effect of our modules: task-specific semantic embedding (\emph{TSE}), cross-domain self-training (\emph{CS}) and cross-attention in \emph{TSE} (\emph{catt}). 
The accuracies on the four target domains are reported in Table \ref{tab:ablation module}. As seen, our method achieve the best performance when three modules are all used. The performance of the single \emph{CS} is the worst that shows that local features cannot align the distributions of the two domains, thus affecting cross-domain self-training. The module \emph{TSE} is introduced into four combinations, all improving the performance, which validates the efficacy of our task-specific semantic features for FS-UDA again. Also, the addition of cross-attention into \emph{TSE} will further improve the performance, which can help discover more semantics from previous tasks.

\textbf{Ablation study of different losses.} We conduct various experiments on \emph{DomainNet} to further evaluate the effect of different losses in Eq. (\ref{total loss}). Besides the classification loss ($\mathcal{L}_{cls}$), we combine the remaining three loss terms: 1) semantic features alignment loss ($\mathcal{L}_{sfa}$), 2) similarity pattern alignment loss ($\mathcal{L}_{spa}$), and 3) class matching loss ($\mathcal{L}_{clm}$). We evaluate 5-way 1-shot  FS-UDA tasks from \emph{sketch} to the other four domains, respectively, and their accuracies are reported in Table \ref{tab:ablation loss}. As observed, the more the number of loss terms involved, the higher the accuracy. The combination of all the three losses is the best. For the single loss, both $\mathcal{L}_{sfa}$ and $\mathcal{L}_{clm}$ perform better than $\mathcal{L}_{spa}$, 
and their combination is also considerably better than the other paired combinations, showing the efficacy of semantic feature domain alignment and class matching in target domain. Based on the above, adding $\mathcal{L}_{spa}$ further improves the performance, indicating positive effect of aligning the similarity patterns.

\begin{table}
\caption{Ablation study ($\%$) of the three losses designed in TSECS, where the FS-UDA tasks are evaluated from a domain (\emph{sketch}) to the other four domains in \emph{DomainNet}.}
\label{tab:ablation loss}
\vspace{-0.4cm}
\centering
\scriptsize
\setlength{\tabcolsep}{8pt}
\renewcommand{\arraystretch}{0.95}
\begin{tabular}{ccc|cccc}
\toprule[1.5pt]
\multicolumn{3}{c|}{\textbf{Components}} & \multicolumn{4}{c}{\textbf{Target Domains}} \\
\hline
$\mathcal{L}_{sfa}$ & $\mathcal{L}_{spa}$ & $\mathcal{L}_{clm}$ & \emph{cli} & \emph{rel} & \emph{qdr} & \emph{pnt} \\
\hline
$\checkmark$ & & & 66.67 & 58.84 & 56.91 & 43.28 \\
& $\checkmark$ & & 64.28 & 57.32 & 52.11 & 42.46 \\
& & $\checkmark$ & 66.83 & 58.29 & 56.51 & 44.25 \\
$\checkmark$ & \checkmark &  & 66.64 & 62.64 & 57.41 & 53.40\\
$\checkmark$ & & \checkmark & 68.04 & 63.98 & 59.13 & 55.39\\
& $\checkmark$ & \checkmark & 67.61 & 62.47 & 53.07 & 54.14\\
 $\checkmark$ & $\checkmark$ & \checkmark & \textbf{69.45} & \textbf{65.00} & \textbf{62.25}
 & \textbf{56.51}\\
\bottomrule[1.5pt]
\end{tabular}
\vspace{-0.45cm}
\end{table}

\begin{figure}
    \centering
    \includegraphics[width=\columnwidth]{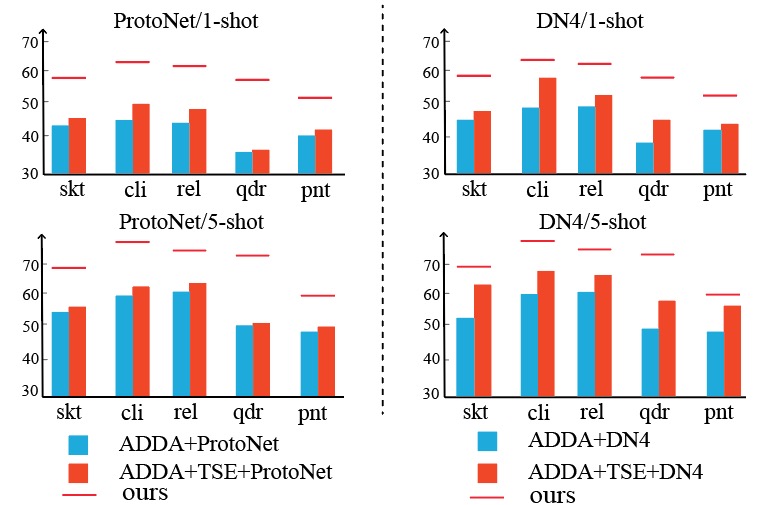}
    \vspace{-0.6cm}
    \caption{Comparison of introducing our TSE module or not into two FSL methods with ADDA \cite{8099799} combined, \emph{i.e.,} ADDA+ProtoNet and ADDA+DN4.}
    \label{semantic_exp}
    \vspace{-0.3cm}
\end{figure}

\textbf{Evaluation on the effect of our task-specific semantic embedding module on two FSL methods with ADDA \cite{8099799} combined.} 
Compared with ADDA+DN4 and ADDA+ProtoNet, we add our semantic embedding module (\emph{TSE}) with the loss $\mathcal{L}_{sfa}$ into their feature embedding models, and test them on 3000 new 5-way 1/5-shot FS-UDA tasks.  
For simplification and clarification, we calculate the averaged accuracies from every domain to the other four domains and show them in Fig. \ref{semantic_exp}. As seen, the methods using \emph{TSE} generally perform better than that without it, which validates that the semantic embedding in \emph{TSE} could generate more discriminative semantic features for classification than original local features. In addition, the performances of these methods are still far from our method because using ADDA is insufficient to align the domains and could result in class mismatching, but our method can effectively solve it by cross-domain self-training.

\begin{table}[h]
\caption{Evaluation ($\%$) of dataset generalization for 5-way 1-shot FS-UDA tasks between domains \emph{real} and \emph{sketch}, performing episodic training on \emph{DomainNet} and testing on expanded dataset \emph{miniImageNet}.} 
\label{tab:general}
\vspace{-0.4cm}
\centering
\scriptsize
\setlength{\tabcolsep}{14pt}
\renewcommand{\arraystretch}{0.95}
\begin{tabular}{c|c|c}
\toprule[1.5pt]
\textbf{Methods} & $skt \rightarrow rel$ & $rel \rightarrow skt$\\
\hline
\textbf{ADDA+DN4} & $44.01\pm0.87$ & $40.61\pm0.90$ \\
\textbf{ADDA+DeepEMD} & $46.14\pm0.82$ & $45.91\pm0.77$ \\
\textbf{IMSE} & $48.78\pm0.78$ & $48.52\pm0.81$ \\
\textbf{TSECS (ours)} & $\textbf{53.33}\pm1.08$ & $\textbf{49.83}\pm0.96$ \\
\bottomrule[1.5pt]
\end{tabular}
\vspace{-0.4cm}
\end{table}

\begin{figure}[htbp]
    \centering
    \includegraphics[width=0.96\columnwidth]{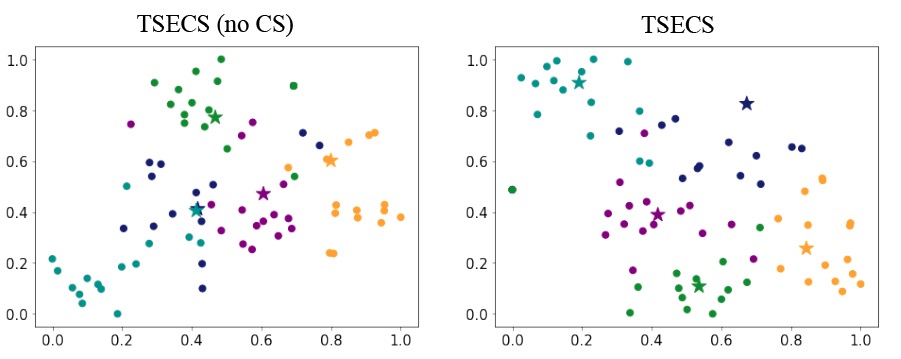}
    \vspace{-0.35cm}
    \caption{The \emph{tSNE} visualization of our TSECS using cross-domain self-training or not for a 5-way 5-shot FS-UDA task from \emph{sketch} to \emph{clipart}. The samples with different colors belong to different classes, and the stars in the left and right figures represent the class centroids of support set and selected target domain query samples, respectively.}
    \label{fig:visual}
    \vspace{-0.45cm}
\end{figure}

\textbf{Evaluation of dataset generalization.} We evaluate the generalization of our model trained on \emph{DomainNet} to adapt to a substantially different dataset \emph{miniImageNet}. We modify \emph{miniImageNet} by transferring a half of real images ($rel$) into sketch images ($skt$) by MUNIT \cite{10.1007/978-3-030-01219-9_11} to produce two domains for FS-UDA. We compare our method with ADDA+DN4, ADDA+DeepEMD and IMSE for 5-way 1-shot FS-UDA tasks for $rel\leftrightarrow skt$. The results are shown as Table \ref{tab:general}. As observed, our method outperforms other methods, specially for \emph{ske} $\rightarrow$ \emph{rel}. 
For \emph{rel} $\rightarrow$ \emph{skt}, our method is slightly better than IMSE, because the style of \emph{sketch} images in \emph{miniImageNet} is relatively different from that in \emph{DomainNet}, which could effect the learned semantic features. 

\textbf{Visualization of our method using cross-domain self-training or not}. We illustrate the \emph{tSNE} results of a 5-way 5-shot FS-UDA task from \emph{sketch} to \emph{clipart} in Fig. \ref{fig:visual}. Note that the class prototypes in the left subfigure belong to the support set in source domain, while those in the right subfigure are generated by `confidence' samples in target domain. It is obvious that two class prototypes in the left subfigure are fully overlapped so that many samples could not be correctly classified. In contrast, the right subfigure has the better class prototypes, and samples from different classes are more distinguishable. This shows the efficacy of our cross-domain self-training that finds `confidence' samples to train the target domain classifier and uses class matching loss $\mathcal{L}_{clm}$ to shorten the distance of samples of the same class.

\section{Conclusion}
In this paper, we propose a novel method TSECS for FS-UDA. We extract high-level semantic features than local features to measure the similarity of query images in target domain to support classes in source domain. Moreover, we design cross-domain self-training to train a target domain classifier. In addition, asymmetrical KL-divergence is used to align the semantic features between domains. Extensive experiments on \emph{DomainNet} show the efficacy of our TSECS, significantly improving the performance for FS-UDA.

\section{Acknowledgments}

\noindent Wanqi Yang and Ming Yang are supported by National Natural
Science Foundation of China (Grant Nos. 62076135, 62276138, 61876087). Lei
Wang is supported by an Australian Research Council Discovery
Project (No. DP200101289) funded by the Australian Government.

\bibliography{aaai23}

\begin{thebibliography}{29}
\providecommand{\natexlab}[1]{#1}

\bibitem[{Bertinetto et~al.(2019)Bertinetto, Henriques, Torr, and
  Vedaldi}]{bertinetto2018metalearning}
Bertinetto, L.; Henriques, J.~F.; Torr, P.; and Vedaldi, A. 2019.
\newblock Meta-learning with differentiable closed-form solvers.
\newblock In \emph{International Conference on Learning Representations}, 1--8.

\bibitem[{Chen, Fan, and Panda(2021)}]{9711309}
Chen, C.-F.~R.; Fan, Q.; and Panda, R. 2021.
\newblock CrossViT: Cross-Attention Multi-Scale Vision Transformer for Image
  Classification.
\newblock In \emph{2021 IEEE/CVF International Conference on Computer Vision
  (ICCV)}, 347--356.

\bibitem[{Chen et~al.(2019)Chen, Liu, Kira, Wang, and Huang}]{chen2018a}
Chen, W.-Y.; Liu, Y.-C.; Kira, Z.; Wang, Y.-C.~F.; and Huang, J.-B. 2019.
\newblock A Closer Look at Few-shot Classification.
\newblock In \emph{International Conference on Learning Representations},
  1--16.

\bibitem[{Finn, Abbeel, and Levine(2017)}]{pmlr-v70-finn17a}
Finn, C.; Abbeel, P.; and Levine, S. 2017.
\newblock Model-Agnostic Meta-Learning for Fast Adaptation of Deep Networks.
\newblock In Precup, D.; and Teh, Y.~W., eds., \emph{Proceedings of the 34th
  International Conference on Machine Learning}, volume~70 of \emph{Proceedings
  of Machine Learning Research}, 1126--1135.

\bibitem[{Ganin et~al.(2016)Ganin, Ustinova, Ajakan, Germain, Larochelle,
  Laviolette, March, and Lempitsky}]{JMLR:v17:15-239}
Ganin, Y.; Ustinova, E.; Ajakan, H.; Germain, P.; Larochelle, H.; Laviolette,
  F.; March, M.; and Lempitsky, V. 2016.
\newblock Domain-Adversarial Training of Neural Networks.
\newblock \emph{Journal of Machine Learning Research}, 17(59): 1--35.

\bibitem[{Gidaris et~al.(2020)Gidaris, Bursuc, Komodakis, Perez, and
  Cord}]{Gidaris_2020_CVPR}
Gidaris, S.; Bursuc, A.; Komodakis, N.; Perez, P.; and Cord, M. 2020.
\newblock Learning Representations by Predicting Bags of Visual Words.
\newblock In \emph{Proceedings of the IEEE/CVF Conference on Computer Vision
  and Pattern Recognition (CVPR)}, 6926--6936.

\bibitem[{Huang et~al.(2021)Huang, Yang, Wang, Zhou, and
  Yang}]{10.1145/3474085.3475232}
Huang, S.; Yang, W.; Wang, L.; Zhou, L.; and Yang, M. 2021.
\newblock Few-Shot Unsupervised Domain Adaptation with Image-to-Class Sparse
  Similarity Encoding.
\newblock In \emph{Proceedings of the 29th ACM International Conference on
  Multimedia}, MM '21, 677--685. New York, NY, USA: Association for Computing
  Machinery.
\newblock ISBN 9781450386517.

\bibitem[{Huang et~al.(2018)Huang, Liu, Belongie, and
  Kautz}]{10.1007/978-3-030-01219-9_11}
Huang, X.; Liu, M.-Y.; Belongie, S.; and Kautz, J. 2018.
\newblock Multimodal Unsupervised Image-to-Image Translation.
\newblock In Ferrari, V.; Hebert, M.; Sminchisescu, C.; and Weiss, Y., eds.,
  \emph{Computer Vision -- ECCV 2018}, 179--196. Cham: Springer International
  Publishing.
\newblock ISBN 978-3-030-01219-9.

\bibitem[{Kim et~al.(2021)Kim, Saito, Oh, Plummer, Sclaroff, and
  Saenko}]{9710187}
Kim, D.; Saito, K.; Oh, T.-H.; Plummer, B.~A.; Sclaroff, S.; and Saenko, K.
  2021.
\newblock CDS: Cross-Domain Self-supervised Pre-training.
\newblock In \emph{2021 IEEE/CVF International Conference on Computer Vision
  (ICCV)}, 9103--9112.

\bibitem[{Li et~al.(2020)Li, Wang, Huo, Shi, Gao, and Luo}]{ijcai2020-409}
Li, W.; Wang, L.; Huo, J.; Shi, Y.; Gao, Y.; and Luo, J. 2020.
\newblock Asymmetric Distribution Measure for Few-shot Learning.
\newblock In Bessiere, C., ed., \emph{Proceedings of the Twenty-Ninth
  International Joint Conference on Artificial Intelligence, {IJCAI-20}},
  2957--2963. International Joint Conferences on Artificial Intelligence
  Organization.
\newblock Main track.

\bibitem[{Li et~al.(2019)Li, Wang, Xu, Huo, Gao, and Luo}]{Li_2019_CVPR}
Li, W.; Wang, L.; Xu, J.; Huo, J.; Gao, Y.; and Luo, J. 2019.
\newblock Revisiting Local Descriptor Based Image-To-Class Measure for Few-Shot
  Learning.
\newblock In \emph{Proceedings of the IEEE/CVF Conference on Computer Vision
  and Pattern Recognition (CVPR)}, 7260--7268.

\bibitem[{Long et~al.(2015)Long, Cao, Wang, and Jordan}]{pmlr-v37-long15}
Long, M.; Cao, Y.; Wang, J.; and Jordan, M. 2015.
\newblock Learning Transferable Features with Deep Adaptation Networks.
\newblock In Bach, F.; and Blei, D., eds., \emph{Proceedings of the 32nd
  International Conference on Machine Learning}, volume~37 of \emph{Proceedings
  of Machine Learning Research}, 97--105. Lille, France: PMLR.

\bibitem[{Luo et~al.(2020)Luo, Liu, Guan, Yu, and Yang}]{NEURIPS2020_ed265bc9}
Luo, Y.; Liu, P.; Guan, T.; Yu, J.; and Yang, Y. 2020.
\newblock Adversarial Style Mining for One-Shot Unsupervised Domain Adaptation.
\newblock In Larochelle, H.; Ranzato, M.; Hadsell, R.; Balcan, M.; and Lin, H.,
  eds., \emph{Advances in Neural Information Processing Systems}, volume~33,
  20612--20623. Curran Associates, Inc.

\bibitem[{Peng et~al.(2019)Peng, Bai, Xia, Huang, Saenko, and Wang}]{9010750}
Peng, X.; Bai, Q.; Xia, X.; Huang, Z.; Saenko, K.; and Wang, B. 2019.
\newblock Moment Matching for Multi-Source Domain Adaptation.
\newblock In \emph{2019 IEEE/CVF International Conference on Computer Vision
  (ICCV)}, 1406--1415.

\bibitem[{Ravi and Larochelle(2017)}]{ravi2017optimization}
Ravi, S.; and Larochelle, H. 2017.
\newblock Optimization as a Model for Few-Shot Learning.
\newblock In \emph{International Conference on Learning Representations}.

\bibitem[{Roy et~al.(2019)Roy, Siarohin, Sangineto, Bulo, Sebe, and
  Ricci}]{Roy_2019_CVPR}
Roy, S.; Siarohin, A.; Sangineto, E.; Bulo, S.~R.; Sebe, N.; and Ricci, E.
  2019.
\newblock Unsupervised Domain Adaptation Using Feature-Whitening and Consensus
  Loss.
\newblock In \emph{Proceedings of the IEEE/CVF Conference on Computer Vision
  and Pattern Recognition (CVPR)}, 9471--9480.

\bibitem[{Saito et~al.(2018)Saito, Watanabe, Ushiku, and
  Harada}]{Saito_2018_CVPR}
Saito, K.; Watanabe, K.; Ushiku, Y.; and Harada, T. 2018.
\newblock Maximum Classifier Discrepancy for Unsupervised Domain Adaptation.
\newblock In \emph{Proceedings of the IEEE Conference on Computer Vision and
  Pattern Recognition (CVPR)}, 3723--3732.

\bibitem[{Snell, Swersky, and Zemel(2017)}]{NIPS2017_cb8da676}
Snell, J.; Swersky, K.; and Zemel, R. 2017.
\newblock Prototypical Networks for Few-shot Learning.
\newblock In Guyon, I.; Luxburg, U.~V.; Bengio, S.; Wallach, H.; Fergus, R.;
  Vishwanathan, S.; and Garnett, R., eds., \emph{Advances in Neural Information
  Processing Systems}, volume~30, 4077--4087. Curran Associates, Inc.

\bibitem[{Tang, Chen, and Jia(2020)}]{Tang_2020_CVPR}
Tang, H.; Chen, K.; and Jia, K. 2020.
\newblock Unsupervised Domain Adaptation via Structurally Regularized Deep
  Clustering.
\newblock In \emph{Proceedings of the IEEE/CVF Conference on Computer Vision
  and Pattern Recognition (CVPR)}.

\bibitem[{Tseng et~al.(2020)Tseng, Lee, Huang, and
  Yang}]{Tseng2020Cross-Domain}
Tseng, H.-Y.; Lee, H.-Y.; Huang, J.-B.; and Yang, M.-H. 2020.
\newblock Cross-Domain Few-Shot Classification via Learned Feature-Wise
  Transformation.
\newblock In \emph{International Conference on Learning Representations}.

\bibitem[{Tzeng et~al.(2017)Tzeng, Hoffman, Saenko, and Darrell}]{8099799}
Tzeng, E.; Hoffman, J.; Saenko, K.; and Darrell, T. 2017.
\newblock Adversarial Discriminative Domain Adaptation.
\newblock In \emph{2017 IEEE Conference on Computer Vision and Pattern
  Recognition (CVPR)}, 2962--2971.

\bibitem[{Tzeng et~al.(2014)Tzeng, Hoffman, Zhang, Saenko, and
  Darrell}]{DBLP:journals/corr/TzengHZSD14}
Tzeng, E.; Hoffman, J.; Zhang, N.; Saenko, K.; and Darrell, T. 2014.
\newblock Deep Domain Confusion: Maximizing for Domain Invariance.
\newblock \emph{CoRR}, abs/1412.3474: 1--9.

\bibitem[{Vinyals et~al.(2016)Vinyals, Blundell, Lillicrap, kavukcuoglu, and
  Wierstra}]{NIPS2016_90e13578}
Vinyals, O.; Blundell, C.; Lillicrap, T.; kavukcuoglu, k.; and Wierstra, D.
  2016.
\newblock Matching Networks for One Shot Learning.
\newblock In Lee, D.; Sugiyama, M.; Luxburg, U.; Guyon, I.; and Garnett, R.,
  eds., \emph{Advances in Neural Information Processing Systems}, volume~29,
  3630--3638. Curran Associates, Inc.

\bibitem[{Yang et~al.(2022)Yang, Yang, Huang, Wang, and Yang}]{Yang}
Yang, W.; Yang, C.; Huang, S.; Wang, L.; and Yang, M. 2022.
\newblock Few-shot Unsupervised Domain Adaptation via Meta Learning.
\newblock In \emph{IEEE International Conference on Multimedia and Expo
  (ICME)}.

\bibitem[{Ye et~al.(2020)Ye, Hu, Zhan, and Sha}]{Ye_2020_CVPR}
Ye, H.-J.; Hu, H.; Zhan, D.-C.; and Sha, F. 2020.
\newblock Few-Shot Learning via Embedding Adaptation With Set-to-Set Functions.
\newblock In \emph{Proceedings of the IEEE/CVF Conference on Computer Vision
  and Pattern Recognition (CVPR)}, 8805--8814.

\bibitem[{Yue et~al.(2021)Yue, Zheng, Zhang, Gao, Darrell, Keutzer, and
  Vincentelli}]{Yue_2021_CVPR}
Yue, X.; Zheng, Z.; Zhang, S.; Gao, Y.; Darrell, T.; Keutzer, K.; and
  Vincentelli, A.~S. 2021.
\newblock Prototypical Cross-Domain Self-Supervised Learning for Few-Shot
  Unsupervised Domain Adaptation.
\newblock In \emph{Proceedings of the IEEE/CVF Conference on Computer Vision
  and Pattern Recognition (CVPR)}, 13834--13844.

\bibitem[{Zhang et~al.(2020)Zhang, Cai, Lin, and Shen}]{Zhang_2020_CVPR}
Zhang, C.; Cai, Y.; Lin, G.; and Shen, C. 2020.
\newblock DeepEMD: Few-Shot Image Classification With Differentiable Earth
  Mover's Distance and Structured Classifiers.
\newblock In \emph{Proceedings of the IEEE/CVF Conference on Computer Vision
  and Pattern Recognition (CVPR)}, 12200--12210.

\bibitem[{Zou et~al.(2019)Zou, Yu, Liu, Kumar, and Wang}]{9010413}
Zou, Y.; Yu, Z.; Liu, X.; Kumar, B. V. K.~V.; and Wang, J. 2019.
\newblock Confidence Regularized Self-Training.
\newblock In \emph{2019 IEEE/CVF International Conference on Computer Vision
  (ICCV)}, 5981--5990.

\bibitem[{Zou et~al.(2018)Zou, Yu, Vijaya~Kumar, and
  Wang}]{10.1007/978-3-030-01219-9_18}
Zou, Y.; Yu, Z.; Vijaya~Kumar, B. V.~K.; and Wang, J. 2018.
\newblock Unsupervised Domain Adaptation for Semantic Segmentation via
  Class-Balanced Self-training.
\newblock In Ferrari, V.; Hebert, M.; Sminchisescu, C.; and Weiss, Y., eds.,
  \emph{Computer Vision -- ECCV 2018}, 297--313. Cham: Springer International
  Publishing.
\newblock ISBN 978-3-030-01219-9.

\end{thebibliography}

\end{document}